\pdfoutput=1
\documentclass[11pt]{article}

\usepackage{authblk}
\usepackage{EMNLP2023}

\usepackage{subfigure}
\usepackage{algorithm}
\usepackage{algorithmic}
\usepackage{times}
\usepackage{latexsym}
\usepackage{graphicx}
\usepackage{adjustbox}
\usepackage{multirow}
\usepackage[T1]{fontenc}
\usepackage[utf8]{inputenc}
\usepackage{amsfonts}
\usepackage{microtype}
\usepackage{inconsolata}

\newcommand\blfootnote[1]{%
  \begingroup
  \renewcommand\thefootnote{}\footnote{#1}%
  \addtocounter{footnote}{-1}%
  \endgroup
}

\title{SLoRA: Federated Parameter Efficient Fine-Tuning of Language Models}

\author{Sara Babakniya$^{* 1}$, Ahmed Roushdy Elkordy$^{* 1}$, Yahya H. Ezzeldin$^1$, \\ Qingfeng Liu$^2$, Kee-Bong Song$^2$, Mostafa El-Khamy$^2$, Salman Avestimehr$^1$}
\affil{$^1$ University of Southern California, $^2$ SoC R$\&$D Samsung Semiconductor Inc.}

\begin{document}
\maketitle
\blfootnote{$^\ast$ Equal contribution (alphabetical order on the last name).}
\begin{abstract}
Transfer learning via fine-tuning pre-trained transformer models has gained significant success in delivering state-of-the-art results across various NLP tasks. In the absence of centralized data, Federated Learning (FL) can benefit from distributed and private data of the FL edge clients for fine-tuning. However, due to the limited communication, computation, and storage capabilities of edge devices and the huge sizes of popular transformer models, efficient fine-tuning is crucial to make federated training feasible. This work explores the opportunities and challenges associated with applying parameter efficient fine-tuning (PEFT) methods in different FL settings for language tasks. Specifically, our investigation reveals that as the data across users becomes more diverse, the gap between fully fine-tuning the model and employing PEFT methods widens. To bridge this performance gap, we propose a method called SLoRA, which overcomes the key limitations of LoRA in high heterogeneous data scenarios through a novel data-driven initialization technique. Our experimental results demonstrate that SLoRA achieves performance comparable to full fine-tuning, with significant sparse updates with approximately $\sim 1\%$ density while reducing training time by up to $90\%$.
\end{abstract}

\section{Introduction}
With the popularity of smartphones and personal gadgets, valuable user data is distributed more than ever. This data can help different companies and service providers improve their products and make them more efficient and personalized. However, privacy is a growing and crucial concern that is inevitable to avoid. Users care about the performance of their applications but do not want their private data to be accessed by everyone. Federated Learning (FL) \citep{mcmahan2017communication,konevcny2016federated} is a new paradigm that can solve both problems simultaneously. Users collaboratively train a common model locally using their private dataset and only share their model update with a parameter server that orchestrates the training process over multiple training rounds without sharing their private data. 

Although FL has already proven beneficial in various domains \citep{kairouz2021advances}, such as next-word prediction and healthcare, it still has critical challenges to be deployed on large scales. Here we mainly focus on the efficiency of FL and problems of clients' heterogeneous data distribution for pre-trained language models. 

\begin{figure}[t]
\includegraphics[width=0.7\linewidth]{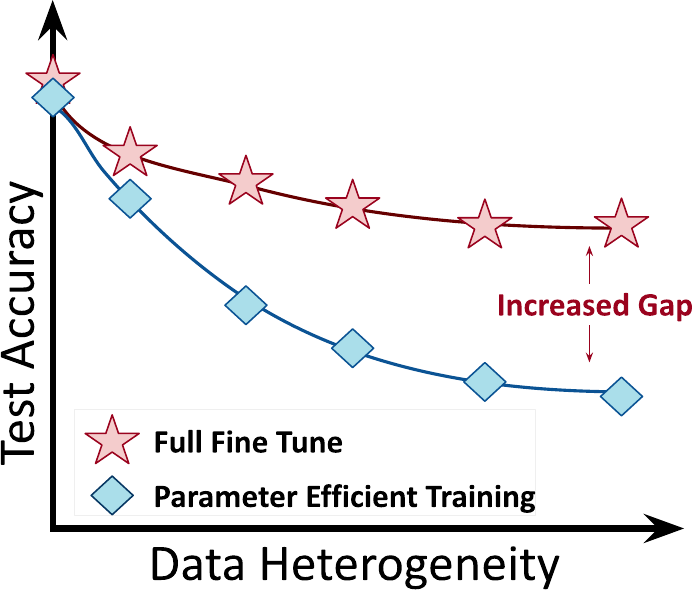}
\centering
   \caption{The impact of client data distribution on the performance of full fine-tuning vs. Parameter efficient Fine-tuning. While heterogeneity has an adverse effect on both of them, parameter efficient methods are more vulnerable and experience more accuracy drop in more heterogeneous settings.}
\label{fig:motivation}
\end{figure}

Large pre-trained language models have proven to perform very well even in the zero-shot setup \citep{brown2020language}. However, as the tasks get more specialized, these models require fine-tuning to enhance their performance on these domain-specific tasks \citep{hu2021lora}. To provide the required privacy while fine-tuning, we can benefit from the FL paradigm. But asking the clients to fine-tune the models and communicate the update has downsides.  

The problem is that fine-tuning can be computationally expensive as it might change the parameters of the entire model. Besides, language models are not used in one application, and clients need to fine-tune them on several tasks. As a result, supporting multi-tasks can also be challenging, especially in memory-constrained scenarios (e.g., for edge devices), because the required memory for the fine-tuned models grows linearly with the number of tasks.

The next concern of shifting the fine-tuning to the client side is its overhead on their already limited resources. Edge devices usually have very little bandwidth (especially up-link) as multiple users share the same resource. Furthermore, both communicating and training can be highly energy-consuming. Therefore, the direct use of FL for NLP tasks may limit its applicability.

Recently, Parameter Efficient Fine Tuning (PEFT) \citep{hu2021lora, li2021prefix, lester2021power} has emerged as an alternative training strategy that does not require fine-tuning of all parameters of the pre-trained model but \textit{only} updates a small portion of the parameters (task-specific parameters) while freezing most of the pre-trained weights of the model to their initial pre-trained values. This approach has been shown to maintain task performance while reducing the parameter budget needed in the centralized setting. Furthermore, if one can successfully design a federated PEFT can also benefit from reduced communication costs.

In this work, we first explore the performance of the existing centralized PEFT method in FL. We observe that the gap between  Full Fine Tuning (FFT) and PEFT increases, that the clients' data distribution gets more heterogeneous (Fig. \ref{fig:motivation}). To this aim, we propose a new algorithm called Primed-LoRA designed for FL where its efficient variant, SLoRA, can achieve first parameter efficiency, second, reduce training and communication cost, and finally closes the gap between PEFT and FFT.  

\section{Related Work}
In the following, we summarize areas in the literature closely related to federated parameter efficient fine-tuning.

\smallskip

\noindent{\bf Parameter Efficient Fine Tuning (PEFT).}
In general,  PEFT methods can be broadly classified into two main categories based on the nature of the tuned parameter. The first category fine-tunes a subset of existing parameters, including the classification head, bias term \citep{zaken2021bitfit}, and sparse subnetworks within the original pre-trained model for each task \citep{guo2020parameter}. The second category is module-based fine-tuning, where an additional set of parameters (e.g., modules) are added for each task. These modules are fine-tuned while freezing the entire pre-trained model. 

Different methods have been proposed depending on the place where the modules are inserted into the model. One class adds bottleneck trainable modules serially to the model components such as adapters ~\citep{houlsby2019parameter} and its variants ~\citep{pfeiffer2022lifting,he2021towards}. Another approach is to introduce modules added in parallel to model parameters such as LoRA~\citep{hu2021lora} and prefix or prompt tuning added in parallel to the attention heads ~\citep{li2021prefix} or embeddings ~\citep{lester2021power}. Recently, there have been several approaches for data-driven PEFT configuration selection for adding adapter modules ~\citep{he2021towards,zhou2023autopeft,wang2022adamix}.

\smallskip

\noindent {\bf PEFT in Federated Learning.} Recent studies \citep{sun2022exploring,zhang2022federated} have investigated the performance of various PEFT methods within the context of Federated Learning (FL) for vision tasks. These studies considered different aspects of federated learning, such as client stability, data distribution, and differential privacy settings. The findings indicated that PEFT could replace FFT without compromising performance while significantly reducing communication costs. 

While the previous works focused on vision and vision-language models, our study differs in several key aspects. Firstly, we specifically study the application of PEFT in the context of language models and additionally examine the effect of data heterogeneity across clients on the performance of PEFT for NLP tasks. Secondly, our work extends beyond benchmarking different PEFT methods in the federated setting to propose an approach that yields comparable performance to FFT even in extreme non-IID settings.

\smallskip

\noindent{\bf Efficient Training in FL.}
Efficient training in FL has been extensively studied in the literature \citep{diao2020heterofl,horvath2021fjord,alam2022fedrolex,niu2022federated,kang2020reliable,li2020federated,li2021fedmask}. Efficient training in FL employs sparse training at different levels. Some approaches only apply sparse training at the client side to update a full-size model retained at the server ~\citep{diao2020heterofl,horvath2021fjord,alam2022fedrolex,niu2022federated}. Other approaches utilize sparse training to optimize a model that is sparse both at the client and server sides~\citep{bibikar2022federated,sfl,qiu2022zerofl,li2021fedmask}.

Although efficient learning and PEFT seem very similar at first glance, given they both share a common goal of reducing the training complexity for the clients. PEFT also focuses on the unique aspect of storage load when considering multiple tasks.

Applying efficient sparse training methods on (pre-trained) large language models typically can only retain good performance with a moderate level of sparsity ~\citep{frantar2023sparsegpt,gordon2020compressing,chen2020lottery}. This can result in a huge storage penalty as sparsification patterns can be heterogeneous across different tasks. On the other hand, PEFT retains the full pre-trained model but applies extremely sparse with ($\sim 1\%$) density updates \citep{hu2021lora,houlsby2019parameter,zaken2021bitfit,pfeiffer2022lifting,he2021towards} to the pre-trained model per task, which allows for substantial storage savings and significant reduction on the communication cost. Importantly, PEFT does this while providing a strong comparable performance to the fully fine-tuned model.

\section{Preliminaries}
\subsection{PEFT Baselines in Centralized Learning}
We investigate Pfieffer, LoRA,  Holusby, and BitFit as state-of-the-art PEFT methods in PEFT. The first three methods add a separate bottleneck module (a down projection dense layer followed by an up projection) with a dimension $r$ to the model. Their difference is where the module is added. Holusby places a  bottleneck module after the multi-head attention and feed-forward block in each Transformer layer. Pfieffer places a bottleneck module only after the feed-forward block in each Transformer layer. In LoRA, the bottleneck module can be parallel to any dense layer in the model. Finally, BitFit is a simple method that only allows fine-tuning the bias terms.

\subsection{Observation: PEFT is challenged when  data distribution gets non-IID}
One of the biggest challenges in FL is the degradation in performance when training in scenarios with heterogeneous client distributions ~\citep{kairouz2021advances}. While this is a well-documented phenomenon reflected in FFT in FL, we observe in this work that the penalty to performance is even more substantial when using PEFT compared to FFT. 
In particular, after benchmarking different models and datasets, we observe that the higher the level of heterogeneity, the more significant the performance gap between FFT and PEFT methods.
Thus, a simple naive adaptation of applying PEFT methods locally in an FL setting can lead to potentially huge performance loss.
 
Our focus in the remainder of the paper is on developing approaches to reduce the gap between FFT and PEFT while efficiently using clients' resources regarding communication and storage loads.
Before proposing our approach in Section~\ref{sec:our_method}, we summarize the approach that shows the greatest promise, which is the SOTA PEFT approach, LoRA.

\begin{figure}[t]
\includegraphics[width=0.6\linewidth]{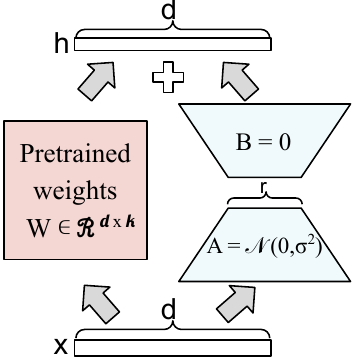}
\centering
   \caption{LoRA Block}
\label{fig:lora}
\end{figure}

\begin{figure*}[t]
  \centering
  \subfigure[20News group dataset on Albert]{\includegraphics[width=0.45\linewidth]{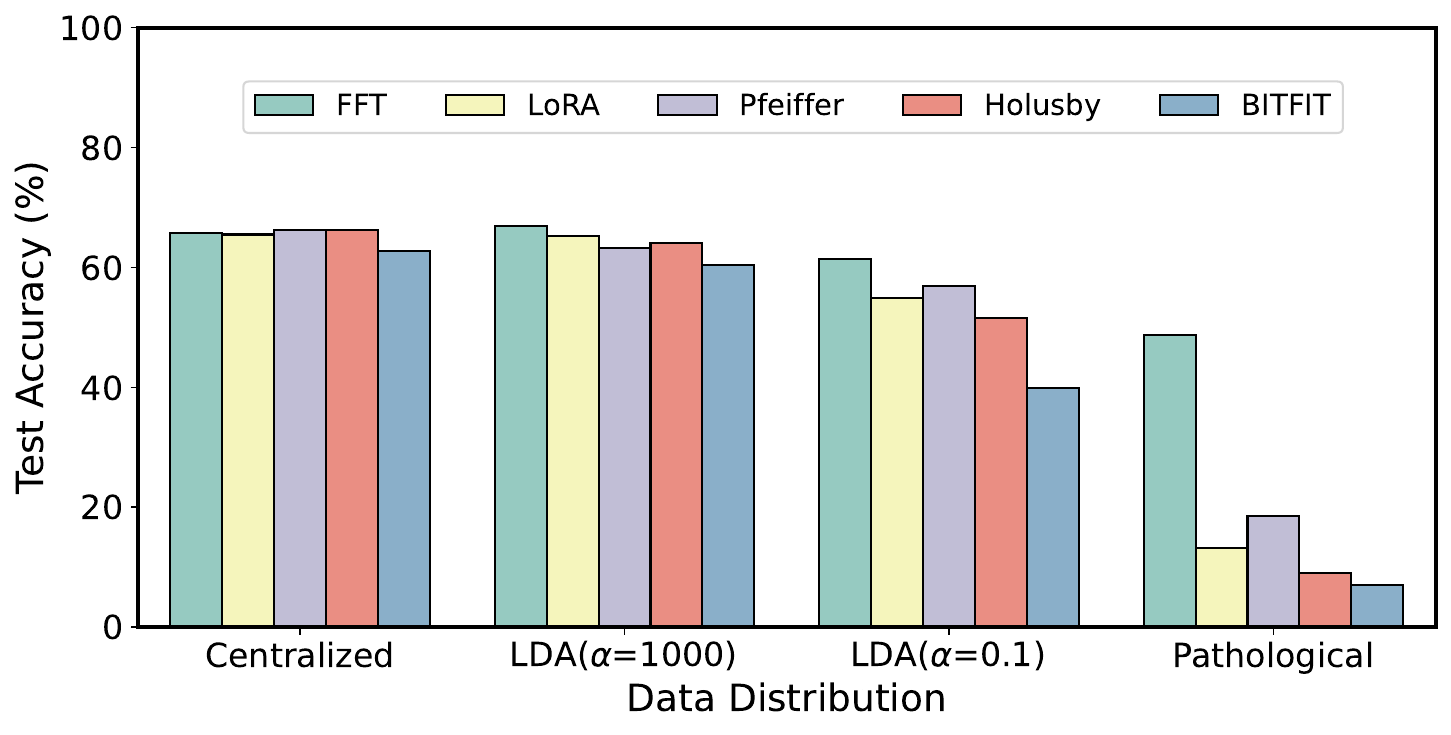}} \quad \quad
  \subfigure[20News group dataset on DistilBERT]{\includegraphics[width=.45\linewidth]{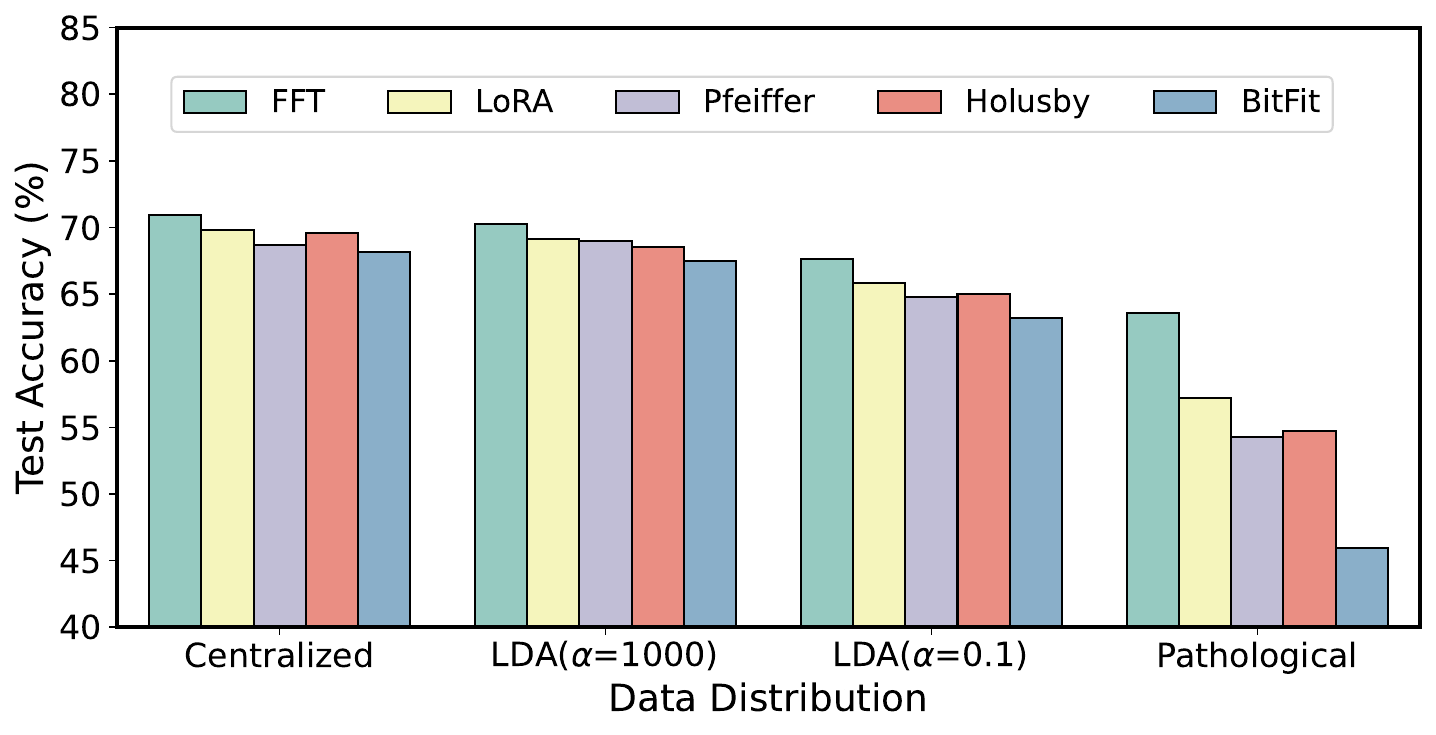}
  }
  \caption{Performance of PEFT methods in the centralized and federated setting with different data distributions for 20News group dataset.}
  \label{fig:data_dist_20news}
\end{figure*}

\subsection{Low-Rank Adaption: LoRA}
We focus on presenting Low-Rank Adaptation (LoRA) since it's the state-of-the-art method for PEFT of large pre-trained language models (LMs), and our proposed algorithm adopts this method. The key idea of LoRA is that instead of fully fine-tuning the pre-trained weight matrix $\mathbf{W}_0 \in \mathbb{R}^{
d\times k}$,  its update is constrained with a low-rank decomposition $\mathbf{W}_0 + \Delta \mathbf{W} = \mathbf{W}_0 + \mathbf{B} \mathbf{A}$, where $\mathbf{B} \in \mathbb{R}^{
d\times r}$, $\mathbf{A} \in \mathbb{R}^{
r\times k}$,   and $ r << \min (k,d)$, and only   $\mathbf{B}$ and  $\mathbf{A}$  are trained while freezing  the pre-trained weight $ \mathbf{W}_0$. 

The way the  LoRA is implemented is shown in Fig. \ref{fig:lora}, where a parallel module of a down projection matrix $\mathbf{B}$ followed by the up projection matrix $\mathbf{A}$ in parallel to original pre-trained weight matrix.    A random Gaussian initialization for  $\mathbf{A}$ and
zero for $\mathbf{B}$ are used to ensure the modified model and the original model are equivalent (e.g.,   $\Delta \mathbf{W} =\mathbf{B} \mathbf{A}$ is zero at the beginning of training).  The modified forward pass after adding the LoRA module is given as follows
\begin{equation} \label{eq}
    \mathbf{h} = \mathbf{W}_0 \mathbf{x} + \frac{
\beta}{r}\mathbf{B} \mathbf{A} \mathbf{x},
\end{equation}
where $r$ is  LoRA rank and $\beta$ 
is a constant in $r$. Therefore, according to (eq. \ref{eq}), the output from the LoRA module is added coordinate-wise to the output of the original model. The scaling $\frac{
\beta}{r}$  can be used  to reduce the need to re-tune hyper-parameters when  varying
$r$. 

\section{Our Proposed approach (Primed-LoRA)}\label{sec:our_method}
In centralized learning, LoRA consistently has shown promising performance in different tasks and closely follows the FFT accuracy. As shown in Fig. \ref{fig:data_dist_20news}, this still holds for federated settings with more homogeneous data distribution (larger $\alpha$). However, in highly non-IID data distribution, LoRA can fail to reach close to the FFT performance or suffer from a slower convergence rate compared to FFT. We hypothesize that one of the reasons behind this is the way that LoRA blocks are initialized (Fig. \ref{fig:lora}). 

As described in~\citet{hu2021lora}, LoRA initializes the $A$ matrix with random independent Gaussian coefficients and $B$ with 0. This initialization works in a centralized setting where the data is ample and concentrated. However, these random and zero initializations in FL can potentially slow down the fine-tuning process.

\noindent {\bf Data-driven priming of LoRA.} 
Based on our hypothesis, picking a better starting point for LoRA might improve its performance. Therefore, we propose a two-stage parameter efficient fine-tuning, Primed-LoRA, based on the LoRA algorithm. 

In Stage 1, the FL clients collaboratively find a mature starting point to prime the LoRA blocks. Then, in Stage 2, we run the LoRA algorithm with our learned initializers from Stage 1. 
In the remainder of the section, we discuss different ways of priming in Primed-LoRA and the properties of the different priming methods.

\subsection{Full fine-tuning for priming LoRA} \label{sec:FFT_LoRA}
A straightforward approach for Stage 1 of Primed-LoRA is to perform a full fine-tuning for a few rounds in Stage 1 of Primed-LoRA and then use SVD matrix decomposition t extract a good initialization for Stage 2. We call this variant of Primed-LoRA as FFT-LoRA or \textbf{FLoRA}.

Formally, we use $\Delta W$ to denote the accumulated change in the model parameters after Stage 1. This weight difference is, conceptually, the same as what we have in each LoRA block, but $\Delta W$ is of size $d \times k$ matrix. We use SVD to arrive at a low-rank approximation of $\Delta W = \mathbf{B} \times \mathbf{A}$, with $\mathbf{B} \in \mathbb{R}^{
d\times r}$, $\mathbf{A} \in \mathbb{R}^{
r\times k}$ that is used to prime LoRA in Stage 2. A description of how we create $\mathbf{A}$ and $\mathbf{B}$ using SVD is delegated to Appendix~\ref{sec:SVD_priming}.

\begin{figure*}
\includegraphics[width=0.95\linewidth]{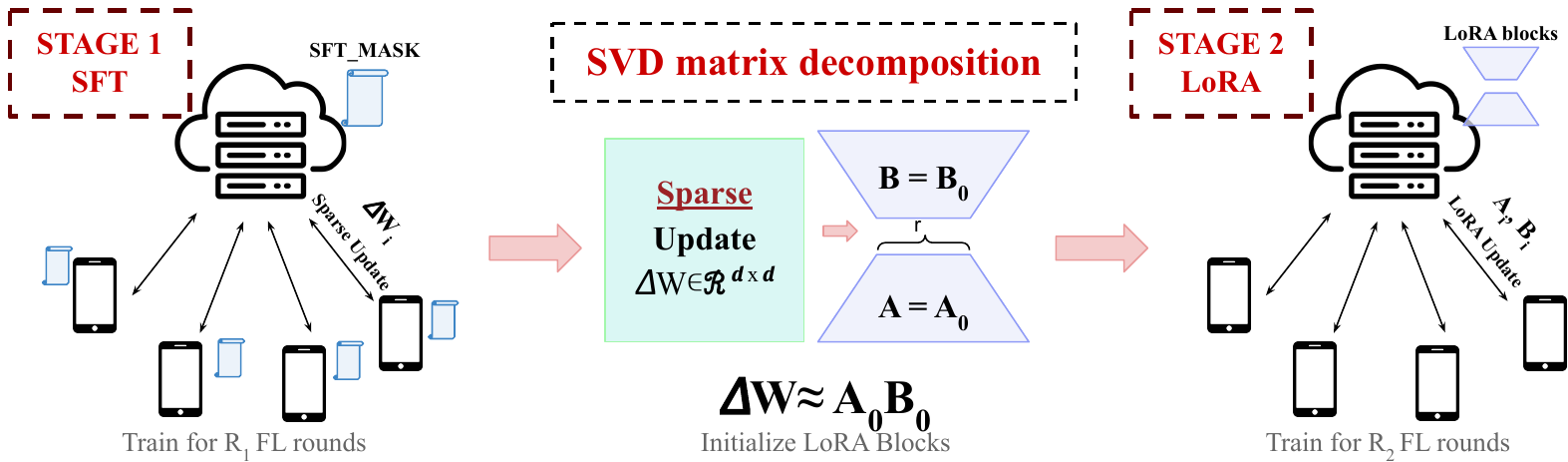}
\centering
   \caption{Overview of SLoRA; First server initializes a mask, and clients only update the parameters in the mask. Then using SVD, the updates are decomposed into LoRA blocks which will be used as an initialization for Stage 2. }
\label{fig:algo}
\end{figure*}

After converting $\Delta W$ into $A$ and $B$, the training goes to the second stage, where clients only update the LoRA blocks and share those parameters with the server. As shown in Fig. ~\ref{fig:albert20-news}, FLoRA can improve the global model's performance, especially with more training rounds in Stage 1. However, as we discuss next, using full fine-tuning to prime LoRA comes at a cost.

\noindent{\bf Cost of FLoRA}
In FLoRA, Stage 1 successfully enhances the performance while achieving parameter efficiency. However, in terms of the training cost, the communication and computation cost of training at this stage is the same as full fine-tuning. In Stage 2, the cost depends on dimensions parameter $r$. 

While FLoRA shows that Primed-LoRA can meet our targets in terms of parameter efficiency, one important question is yet to be answered. How to preserve this performance but reduce the training costs? This is especially important during the training because, in cross-device federated learning, clients have a limited budget.

\begin{figure}[t]
 \includegraphics[width=1\linewidth]{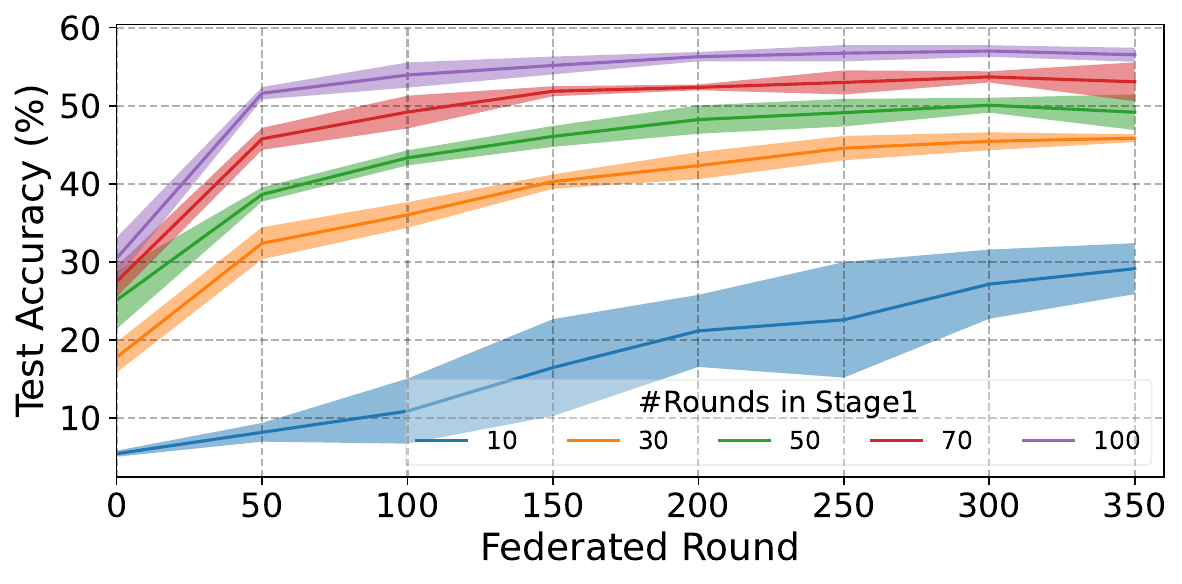}
\centering
   \caption{Impact of the number of federated rounds in stage 1 on the final performance of the model after stage 2 in FLoRA for 20News group dataset on Albert.}
\label{fig:flora_stage1}
\end{figure}

By looking at the cost of FLoRA, we can figure out that there are two parameters involved in this cost; the number of rounds and the communication/computation cost of each round. As a result, one way to decrease the cost in Stage 1 is to reduce the number of FFT rounds. However, as depicted in Fig. \ref{fig:flora_stage1}, the performance of the model at the end of Stage 1 directly impacts the performance of the final model.

Another way to make FLoRA more efficient is to reduce the data processing and update size transmitted from clients. Towards this goal, in the following subsection, we propose \textbf{Primed-LoRA with Sparse Fine-tuning} or SLoRA, where the clients only update a fraction of the parameters in Stage 1 instead of fully fine-tuning the parameters.

\subsection{Sparse Fine-tuning for priming LoRA}\label{sec:SLoRA}
Sparse Fine-Tuning (SFT)~\citep{ansell2021composable,zaken2021bitfit,guo2020parameter} aims to achieve parameter efficiency by sparsifying the updates. In other words, in $W_R = W_0 + \Delta W$, the update ($\Delta W$) is a set of highly sparse matrices that can be stored and transmitted efficiently.

We opt to employ the approach of \citet{ansell2021composable}, but all the works follow similar ideas. In particular, the goal is to find a binary mask such that 1's indicates the position of weights that can change in each round. \citet{ansell2021composable} proposal to generate such mask is to find the \textit{top-K most important} weights based on their contribution in FFT. The weights that change the highest -- from their original pre-trained values -- in FFT are the ones we should keep training in SFT. As a result, the authors add a new warm-up stage where they fine-tune the weight for several rounds to detect such weights.

\noindent \textbf{Sparse Fine-tuning in Stage 1.} \citet{ansell2021composable} is designed for the centralized setting where the data is located in the same location and can be utilized in finding the mask. But, the situation is different in the federated setting. The server cannot fine-tune the model on the current task because it does not have data. Clients can individually train the model and come up with personalized masks. However, the difference in clients' masks increases the density of the aggregated update and reduces parameter efficiency. Alternatively, if the clients want to find the important weights together, they require to do FFT, which is against our goal of reducing the cost and number of FFT communications. 

To solve this problem, we propose the server so generate a random data-independent binary mask with uniform density for all layers at the beginning of training. Then, the clients only train the weights using this mask. Thus, the density of the update from the server and clients does not change, and they can benefit from reduced communication. We want to point out that we choose the mask's density to be higher than the number of parameters in the LoRA blocks. 

\noindent \textbf{Primed-LoRA in Stage 2.} Stage 2 in SLoRA follows the same procedure as discussed in Section~\ref{sec:FFT_LoRA}. After the first Stage 1, we employ SVD to decompose $\Delta W$ into the two components $\mathbf{A}$ and $\mathbf{B}$ used in the LoRA algorithm.
Algorithm~\ref{fig:algo} summarizes the different steps in SLoRA.

\begin{algorithm}[ht!]
    \caption{Overview of Primed LoRA}
    \label{algo:algorithm}
\begin{algorithmic}[1]
\STATE $R_i$: FL rounds in stage $i$, $N$: Total \# clients, $K$: Total \# participant per round, $E$: Local epoch, $W_R$: Model weights in round $R$, $r$: LoRA parameter, $d_{i}$: update density in stage $i$, $Algorithm$ choice between FLoRA and SLoRA

\STATE \textcolor{blue}{\# Stage 1}
\IF{$Algorithm = SLoRA$}
\STATE $SFT\_Mask = generateMask(W_0, d_1)$
\ELSE
\STATE $SFT\_Mask = 1$
\ENDIF
\FOR {$R=1$ {\bfseries to} $R_1$}
    \FOR{$k=1$ {\bfseries to} $K$}    
    \STATE $W^k_{R - 1} = train(W_R, SFT\_Mask, E)$
    \ENDFOR
    \STATE $W_R = aggregate(W^{0,..,k}_{R - 1})$
\ENDFOR
\STATE $\Delta W = W_R - W_0$
\STATE ${[\mathbf{A}, \mathbf{B}]}^0 = SVD(\Delta W, r)$
\STATE \textcolor{blue}{\# Stage 2}
\FOR {$R=1$ {\bfseries to} $R_2$}
    \FOR{$k=1$ {\bfseries to} $K$}    
    \STATE ${[\mathbf{A}, \mathbf{B}]}_R^k = trainLoRA({[\mathbf{A}, \mathbf{B}]}_{R-1}, E)$
    \ENDFOR
    \STATE ${[\mathbf{A}, \mathbf{B}]}_{R} = aggregate({[\mathbf{A}, \mathbf{B}]}_{R-1}^{0,...,K})$
\ENDFOR

\end{algorithmic}
\end{algorithm}

\section{Experiments}\label{sec:experiments}
\subsection{Setting}
\textbf{Models.} Our experiments focus on the following two models: Albert \citet{lan2019albert} and DistilBERT \citet{sanh2019distilbert}.

\noindent\textbf{Datasets.} We show our results for two datasets, News Category and 20News group \citet{lang1995newsweeder}. Since we rely on non-IID label distribution, we focus on classification tasks. The 20News group dataset includes 20 news topics with about 19K data points. We use a 60 $\%$ - 40 $\%$ split for train and test, respectively. The news category is another dataset about news, and it has 15 different labels. This dataset includes about 330K training data, but we only use a randomly sampled $10\%$ for training. Clients only train for one epoch every round, and the batch size is always 32.

\noindent \textbf{Federated Setting.} The total number of clients ($N$) is 100 in all the experiments. The number of participants in every round ($K$) is 20 clients for pathological non-IID and 10 otherwise.

\subsection{Metrics}
We mainly focus on the accuracy of the final global model. Moreover, to compare the different costs of each algorithm, we report their communication cost and training time on our local GPU. We use a single NVIDIA-A100 GPU for each experiment. All the experiments are performed for 5 different seeds, and we report the average of the 3 best results. We used FedAvg \citep{mcmahan2017communication} as our aggregation method, where clients at round $R$ receive $W_R$ from the server. After fine-tuning, clients calculate the difference of the current weight with $W_R$ and send it to the server. Finally, the server average all the updates from all the participants to compute $W_{R+1}$. 

 \begin{figure*}[!h]
  \centering
  \  \subfigure[News Category dataset on Albert]{\includegraphics[width=0.45\linewidth]{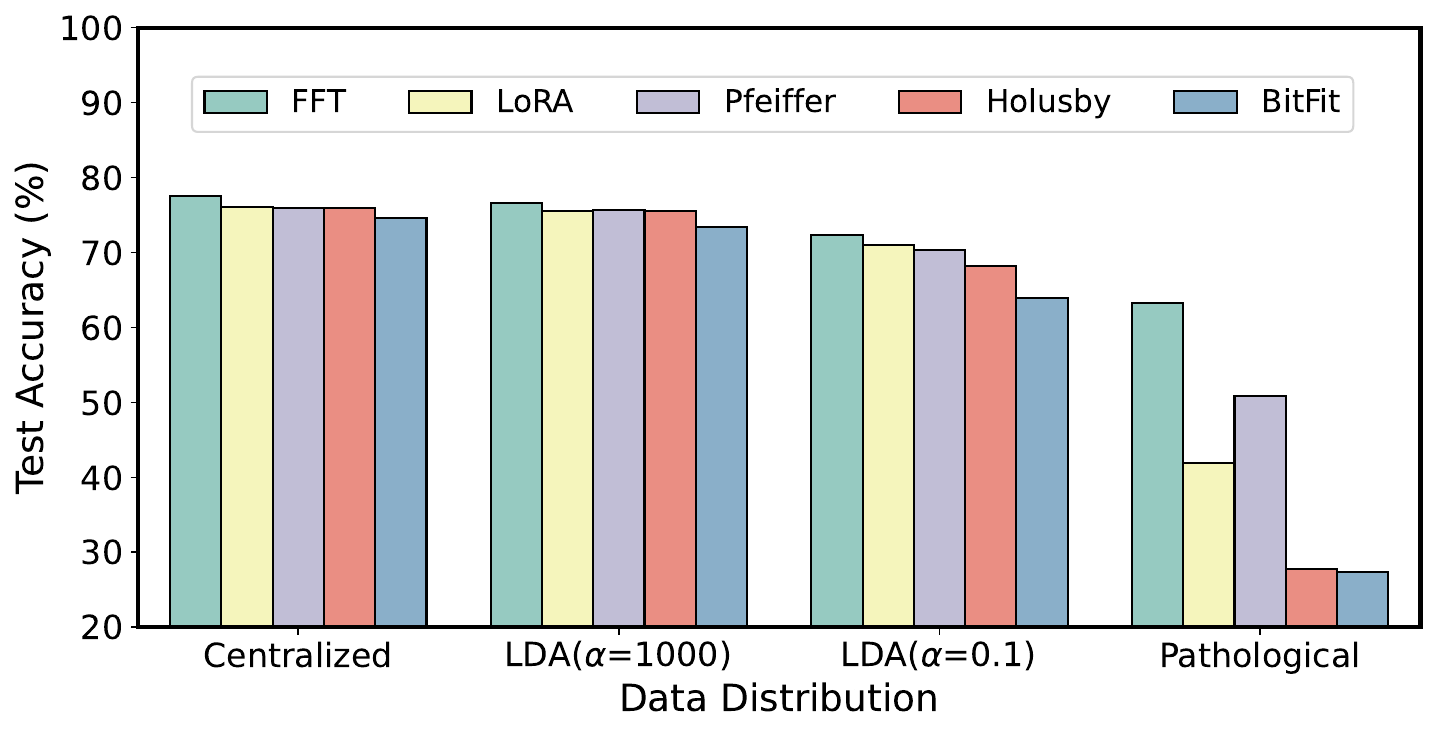}} \quad \quad
    \subfigure[News Category dataset on DistilBERT]{\includegraphics[width=.45\linewidth]{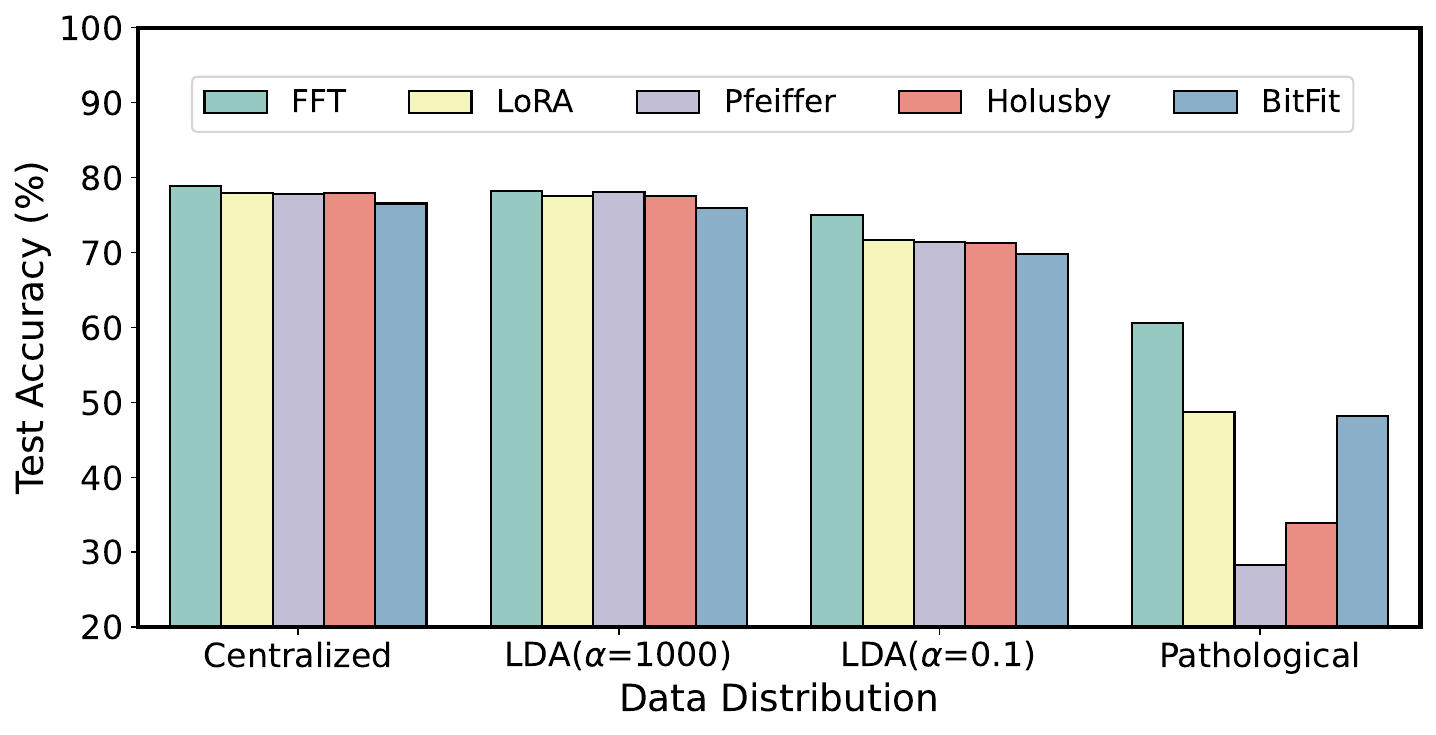}
  }
  \caption{Performance of PEFT methods in the settings and data distributions for News Category dataset.}
  \label{fig:data_dist_category}
 \end{figure*}

\subsection{Baselines}
In the following, we investigate LoRA, Holusby, Pfieffer, and BitFit to understand the impact of data heterogeneity and update size. For data heterogeneity, we used Latent Dirichlet Distribution (LDA) allocation \citep{reddi2020adaptive}, which controls the data distribution of the dataset in each client with $\alpha$ parameter. Small $\alpha$ in this allocation means the data distribution is more heterogeneous. Also, to explore non-IID, we have added pathologically non-IID similar to \citet{mcmahan2017communication}.

\begin{table}[h]
\centering
\caption{$r$ parameter for each setting. Using this parameter, we can calculate the update size of each method.}
  \begin{adjustbox}{width=0.8\linewidth}
\begin{tabular}{c|c|c|c}
\textbf{Method}  & LoRA & Holusby & Pfieffer\\\hline \hline
\textbf{Small} &  $20$  & $38$  & $76$  \\ \hline
\textbf{Large} &  $190$  & $384$ & $768$\\\hline
\end{tabular}
\label{table1:model_param}
\end{adjustbox}
\end{table}

The update size in BitFit is fixed and equal to the total size of bias layers. For other algorithms, update size is controlled by parameter $r$, which indicates the size of the appended module and is summarized in Table \ref{table1:model_param}. Generally, smaller $r$ leads to smaller update sizes, and here we consider two different sizes. The density of the smaller update is approximately similar to BitFit's. To explore the impact of model size, we have included a larger update size with a higher density as well.
 
Also, for the LoRA algorithm, we can add the blocks in all the dense layers, but here we only select the 4 dense layers in the multi-head attention layer, similar to the original paper. LoRA has an extra $\alpha$ parameter which we set to be equal to  $r$.

\section{Evaluation}
\subsection{Data Heterogeneity}
Fig. \ref{fig:data_dist_20news} and \ref{fig:data_dist_category} show the impact of data heterogeneity on the performance of two models, DistilBERT and Albert, trained on 20News and News category datasets. The gap between the PEFT and FFT grows in all four settings by making the data more non-IID. This phenomenon indicates the necessity of a new PEFT algorithm tailored for FL. 

\subsection{Update Size}
Table \ref{table:updatesize} shows the impact of update size for the 20News group dataset on the Albert model on the final performance of the global model. As expected, larger update sizes can help the performance and give a model with higher accuracy. But, the problem with this approach is that now the update size and communication cost of training would also increase, which is not desired.  
\begin{table}[h]
	\addtolength{\tabcolsep}{-3.5pt}
		\caption{Impact of update density of different PEFT methods on the performance for the 20News group dataset on Albert.}
        \vspace{1mm}
		\label{table:updatesize}
		\centering
		\begin{adjustbox}{width=0.9\linewidth}
		\begin{tabular}{c|c|c|c|c}
 \textbf{Data} & \textbf{Density} & \textbf{LoRA} & \textbf{Holusby} & \textbf{Pfeiffer}\\
\textbf{Distribution}& \textbf{($\%$)} & \textbf{($\%$)} & \textbf{$\%$} & \textbf{($\%$)} \\  \hline \hline
\multirow{2}{*}{Centralized}  & $10$ & $65.9$ & $66$ & $66.5$ \\
                              & $1$ & $65.5$ & $66.3$ & $ 66.2$ \\\hline

\multirow{2}{*}{$\alpha=0.1$}  & $10$ & $61.4$ & $53.1$ & $58$ \\
                              & $1$ & $54.9$ & $51.6$ & $56.89$ \\\hline

\multirow{2}{*}{Pathological}  & $10$ & $40.2$ & $38.8$ & $21.1$ \\
                              & $1$ & $9$ & $13.22$  & $18.6$ \\\hline
		\end{tabular}
		\end{adjustbox}
\end{table}
  
\begin{figure*}
  \centering
  \subfigure{\includegraphics[width=0.45\linewidth]{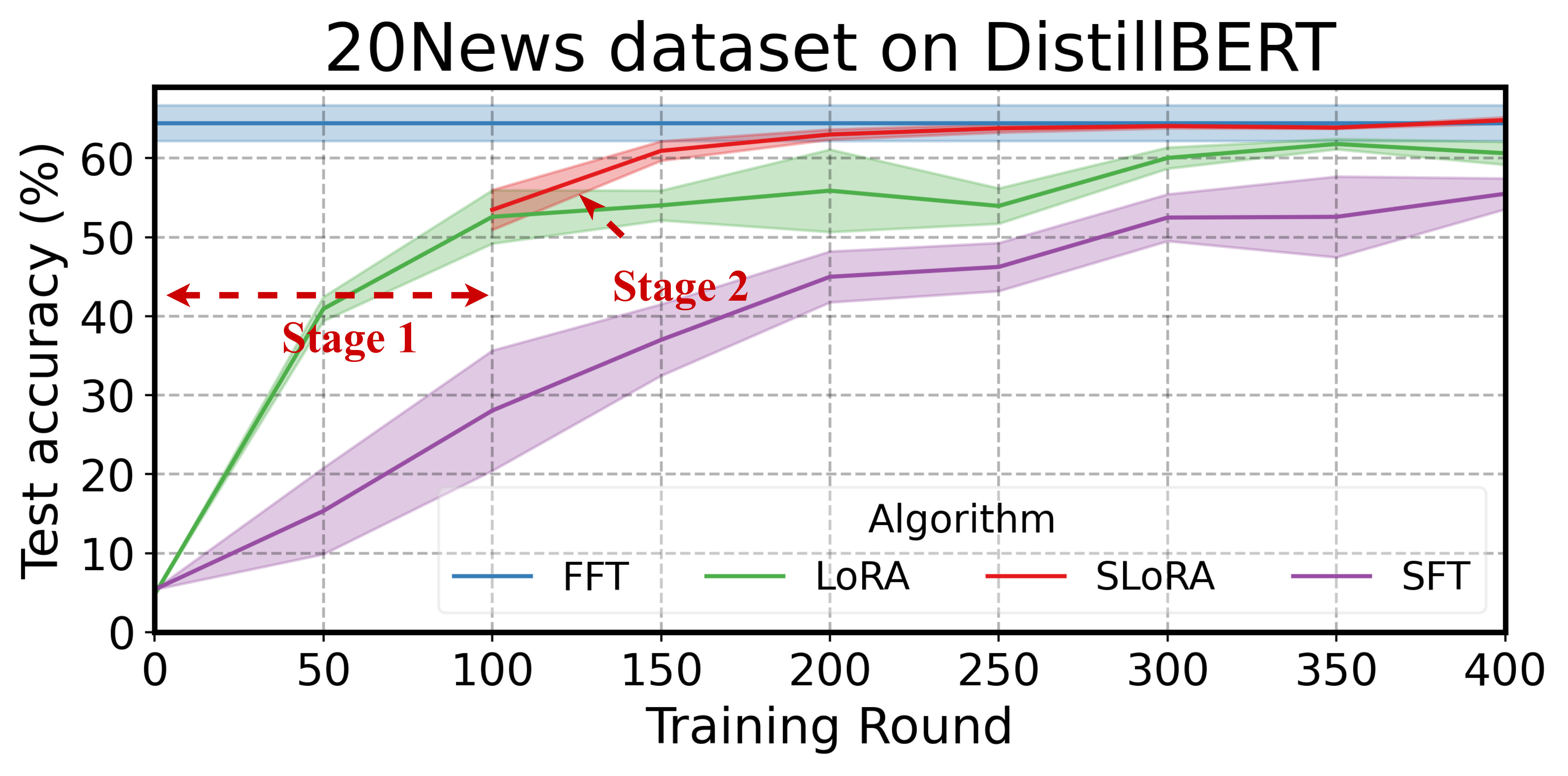} }\quad \quad
  \subfigure{\includegraphics[width=0.45\linewidth]{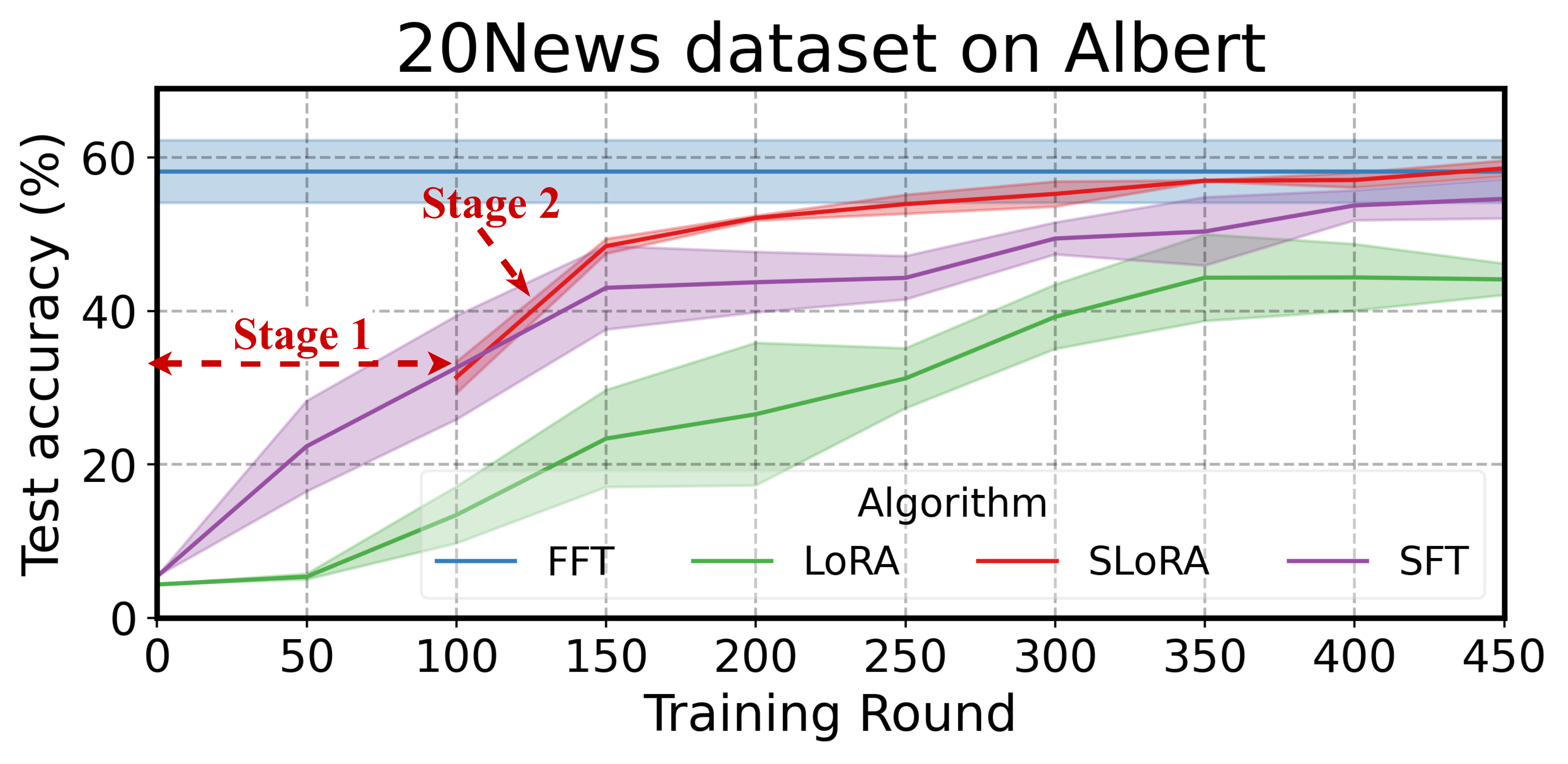}
  }
  \caption{The performance of SLoRA using for 20News group dataset on Albert and DistilBERT.}
  \label{fig:albert20-news}
 \end{figure*}

\begin{table*}[ht]
  \centering
  \caption{The training time (i.e., computation time) of different methods and the number of training rounds with the server for the Albert model on the 20news dataset.}
  \begin{adjustbox}{width=0.95\linewidth}
\begin{tabular}{c|c|c|c|c|c}

\textbf{Method} & \textbf{FFT}  & \textbf{LoRA} & \textbf{SFT}  & \textbf{SLoRA} & \textbf{SVD-decomposition} \\ 
\hline \hline
\textbf{Computation time} (sec/epoch) & $0.39$ & $0.43$ & $2$ & Stage1: $2.1$ - Stage2: $0.43$ &$15.4$ (one-time cost) \\\hline \rule{0pt}{2ex}
\textbf{Total training rounds} & $250$  & $1250$ & $1250$ & $350$      & --- \\\hline
\end{tabular}
\label{table:time_albert}
\end{adjustbox}
\end{table*}
 
\subsection{Performance of SLoRA}
We propose an algorithm called SLoRA which includes two stages. First, clients collaboratively update the model using SFT techniques. Then, the final update is used to generate the initialization for stage 2, which is LoRA. Therefore, we compare our method with \textit{stage 1 without stage 2} and \textit{stage 2 only}, which we call SFT and LoRA, respectively. Through this comparison, we aim to evaluate the extent of improvement achieved by SLoRA in contrast to the individual utilization of each stage. 
 
Here, we only consider pathologically non-IID data distribution as discussed earlier in this setting; PEFT causes a significant drop in its performance. For our proposed method SLoRA, we train the model in Stage 1 using an update sparsity of $10\%$ to have a good model performance at Stage 1 that can be utilized at Stage 2, yet with a minimal cost. In Stage 2, for both SLoRA and FLoRA, we add the LoRA module by utilizing the SVD decomposed model update from Stage 1 with a higher target update sparsity, as discussed in Section \ref{sec:SLoRA}. Specifically, the LoRA modules are added to each dense layer in the model, except for the embedding and classification layers. For the dense layers within the multi-head attention (MHA) block and the feed-forward block, we incorporate the parallel module with an assigned value of $r = 10$.
Additionally, we use $r = 18$ for the pre-classification layer to achieve a better SVD decomposition in the last layer. The size of the added module is $~0.14M$ parameters compared to $~11.7M$ parameters of the original Albert model representing only $1.3\%$ of the original model size.  For SFT, we train the model with a sparsity update of $1.3\%$ to match the same target sparsity update of SLoRA.

\begin{table}[h]
	\addtolength{\tabcolsep}{-3.5pt}
		\caption{Performance and training cost of different algorithm for the 20News group dataset on Albert model.}
        \vspace{1mm}
		\label{table:albert_cost}
		\centering
		\begin{adjustbox}{width=1\linewidth}
		\begin{tabular}{c|c|c|c|c}
& \textbf{$\#$ Trainable} & \textbf{Training} & \textbf{Accuracy} & \textbf{Communication} \\
& \textbf{Parameter} & \textbf{Time (min)} & \textbf{($\%$)} & \textbf{(Gbits)} \\  
			\hline
			\hline
\textbf{FFT} & $11.7M$ & $596.7$& $58.17 \pm 4$ & $174$\\
\hline
\textbf{LoRA} & $\mathbf{0.14} M$ & $49.5$& $56.5 \pm 1.2$& $\mathbf{9.95}$\\
\hline
\textbf{SFT} & $\mathbf{0.14} M$ & $78.4$& $57.6 \pm 0.3$& $\mathbf{9.95}$\\
\hline
\textbf{SLoRA} &$\mathbf{0.14} M$  & $\mathbf{40.4}$ & $\mathbf{58.6 \pm 1}$ & $\mathbf{9.95}$ \\\hline
		\end{tabular}
		\end{adjustbox}
\end{table}

We evaluate the performance of SLoRA for the 20News group dataset on Albert and DistilBERT in Fig. \ref{fig:albert20-news}. As presented in the figure, SLoRA converges to better accuracy and requires smaller training rounds. This is particularly important for low-budget and resource-restricted settings of federated learning. 

Regarding the model training time (e.g., computation time), we report the duration based on a single GPU. Table \ref{table:time_albert} summarizes the average training time per epoch over 10 distinct runs for various methods. We note that the time for SLoRA can be computed using the time for Stages 1 and 2. The SVD decomposition process is executed only once in Stage 2 and can be done efficiently by the server. 

We analyze two aspects of the comparison. Firstly, we evaluate the performance of SLoRA while matching the communication budget across different baselines. In particular, in Stage 1, SLoRA communicates larger models compared to the other PEFT baselines (SFT and LoRA). To ensure a fair comparison, we allow the PEFT baselines to be trained for a longer duration in order to match the same communication budget as SLoRA. Table \ref{table:time_albert} summarizes the number of training rounds for the baselines. Despite the baselines having even higher communication rounds, Fig. \ref{table:albert_cost} demonstrates that SLoRA still achieves comparable accuracy to the fully fine-tuned model, with a slight marginal improvement.

On the other hand, LoRA and SFT show a performance drop of $-1.67$ and $-0.57$ in their maximum accuracy, respectively. It's worth noting that this is achieved with a longer training time compared to SLoRA. Additionally, SLoRA exhibits higher stability than other baselines across different seeds.

In addition to the communication budget, the number of communication rounds is another crucial aspect of federated learning, especially when considering user availability and its impact on computational resource consumption. Therefore, we also compare the performance of SLoRA to the baselines while keeping the number of training rounds fixed, consistent with that of SLoRA. As depicted in Figure \ref{fig:albert20-news}(b), we observe that the performance gap between SLoRA and the baselines (LoRA and SFT) increases to $4\%$ and $14.2\%$, respectively.

\section{Conclusion}
In this work, we have investigated employing PEFT methods for fine-tuning language models in the FL setting to reduce communication and storage costs. We have found that different PEFT methods perform poorly compared to FFT as the diversity of the data increases. To overcome this limitation, we have proposed a novel approach called SLoRA that can maintain the same performance as FFT with minimal communication, time, and storage cost. 

\section*{Limitations}
In this work, we assumed all the clients have similar resource limitations and power. However, in the real world, some users may have more power and can larger update sizes both for training and inference, while others may be more constrained. One interesting future work is extending this work to also consider resource heterogeneity.

\section*{Ethics Statement}
We confirm that this work does not raise any ethical problems. 

\bibliography{anthology}
\bibliographystyle{acl_natbib}

\appendix

\section{Priming LODA from FFT using SVD}
\label{sec:SVD_priming}
Singular Value Decomposition is a matrix decomposition method that helps us to rewrite a $m \times n$ matrix $M$ as a multiplication of three matrices, $M = \mathbf{U} \Sigma \mathbf{V}^T$, where $\mathbf{M} \in \mathbb{R} ^ {m \times n}$, $\mathbf{U} \in \mathbb{R} ^ {m \times m}$, $\Sigma \in \mathbb{R} ^ {m \times n}$, $\mathbf{V} \in \mathbb{R} ^ {n \times n}$ and $\Sigma$ is a rectangular diagonal matrix with descending diagonal terms~\footnote{SVD decomposition is not unique, but we are concerned with the decomposition where the singular values are organized in descending order}.

The accurate decomposition is not parameter efficient and generates matrices with large dimensions. Therefore, instead of the exact decomposition, we use an approximation that preserves most of the information in $\mathbf{M}$ (Our $\Delta W$ of interest throughout the paper). A common way to approximate a $m \times m$ matrix $\mathbf{M} \approx \tilde{\mathbf U}\tilde{\mathbf V}^T$ (where $\tilde{\mathbf U} \in \mathbb{R} ^ {m \times r}$, $\tilde{\mathbf V} \in \mathcal{R} ^ {m \times r}$ and $r \ll m$) is to take the first $r$ columns of $\mathbf{U}$  and $\mathbf{V}$ which are associated with the largest singular values in $\Sigma$, and then constructing $\tilde{\mathbf{U}} = \mathbf{U}_{[1:m,1:r]} \Sigma_{[1:r,1:r]}$ and $\tilde{\mathbf{V}} = \mathbf{V}_{[1:m,1:r]}$. Note that increasing the $r$ value makes this approximation more accurate as it approaches the true SVD but, at the same time, decreases the saving in the parameters.

\section{Performance and cost of 20News group on DistilBERT}
Table \ref{table1:training_time_distilbert} shows the training cost (time) of different baselines and components of SLoRA for the 20News group dataset on the DistilBERT model. As expected, the cost of Stage 1 is similar to SFT, and the cost of Stage 2 is equal to that in LoRA.
\label{sec:appendix_distilbert_20news}
\begin{table}[ht]
  \centering
  \caption{The training time (i.e., computation time) of different methods and the number of training rounds with the server for the Distilbert model on the 20news dataset.}
  \begin{adjustbox}{width=1\linewidth}
\begin{tabular}{c|c|c|c|c|c}
& \textbf{FFT}  & \textbf{LoRA} & \textbf{SFT}  & \textbf{SLoRA} & \textbf{SVD-} \\ 
& & & & & \textbf{decomposition} \\ \hline \hline
\textbf{Computation time} & $0.19$ & $0.20$ & $1.02$ &   Stage 1: $1.1$ & $15.4$ \\ 
(sec/epoch) & & & & Stage 2: $0.20$ & (one-time cost) \\\hline 
\rule{0pt}{2ex}
\textbf{Total training rounds} & $250$  & $1250$ & $1250$ & $300$      & --- \\\hline
\end{tabular}
\label{table1:training_time_distilbert}
\end{adjustbox}
\end{table}

Table \ref{table:performance_d_20} summarizes the performance of different methods for the 20News group dataset on the DistilBERT model. As mentioned earlier, to have a fair comparison, we trained different methods for different rounds of federated learning and only fixed the communication cost. As shown in the table, SLoRA enjoys better performance compared to FFT and still has better training time. 
\begin{table}[h]
\addtolength{\tabcolsep}{-3.5pt}
\caption{Performance and training cost of different algorithm for the 20News group dataset on DistilBERT model.}
\vspace{1mm}
\label{table:performance_d_20}
\centering
\begin{adjustbox}{width=1\linewidth}
\begin{tabular}{c|c|c|c|c}
& \textbf{$\#$ Trainable} & \textbf{Training} & \textbf{Accuracy} & \textbf{Communication} \\
& \textbf{Parameter} & \textbf{Time (min)} & \textbf{($\%$)} & \textbf{(Gbits)} \\  \hline\hline
\textbf{FFT} & $67.0M$ & $3407$& $64.4 \pm 2$ & $997$\\\hline
\textbf{LoRA} & $\mathbf{0.7} M$ & $203$& $63.1 \pm 0.5$& $\mathbf{57.5}$\\\hline
\textbf{SFT} & $\mathbf{0.7} M$ & $220$& $62.3 \pm 0.9$& $\mathbf{57.5}$\\\hline
\textbf{SLoRA} &$\mathbf{0.7} M$  & $\mathbf{186}$ & $\mathbf{64.8 \pm 0.4}$ & $\mathbf{57.5}$ \\\hline
\end{tabular}
\end{adjustbox}
\end{table}

The results for the News category dataset on Albert and DistilBERT are summarized in Table \ref{table:albert_cost-ca} and Table \ref{table:cost-cd}, respectively. As depicted in the tables, the same observations still hold for this dataset as well. 
\begin{table}[h]
	\addtolength{\tabcolsep}{-3.5pt}
		\caption{Performance and training cost of different algorithms for the News category dataset on Albert model.}
        \vspace{1mm}
		\label{table:albert_cost-ca}
		\centering
		\begin{adjustbox}{width=1\linewidth}
		\begin{tabular}{c|c|c|c|c}
& \textbf{$\#$ Trainable} & \textbf{Training} & \textbf{Accuracy} & \textbf{Communication} \\
& \textbf{Parameter} & \textbf{Time (min)} & \textbf{($\%$)} & \textbf{(Gbits)} \\  
			\hline
			\hline
\textbf{FFT} & $11.7M$ & $ 559$& $ 65.2 \pm 0.6$ & $174$\\
\hline
\textbf{LoRA} & $\mathbf{0.14} M$ & $\mathbf{39.5}$& $56.8 \pm 5$& $\mathbf{9.95}$\\
\hline
\textbf{SFT} & $\mathbf{0.14} M$ & $ 140$& $ 55.1 \pm 0.6$& $\mathbf{9.95}$\\
\hline
\textbf{SLoRA} &$\mathbf{0.14} M$  & $ 41.5$ & $\mathbf{62.8 \pm 3}$ & $\mathbf{9.95}$ \\\hline
		\end{tabular}
		\end{adjustbox}
\end{table}

\begin{table}[h]
	\addtolength{\tabcolsep}{-3.5pt}
		\caption{Performance and training cost of different algorithms for the News category dataset on DistilBERT model. We train SLoRA on Stage 2 for only 50 rounds.}
        \vspace{1mm}
		\label{table:cost-cd}
		\centering
		\begin{adjustbox}{width=1\linewidth}
		\begin{tabular}{c|c|c|c|c}
& \textbf{$\#$ Trainable} & \textbf{Training} & \textbf{Accuracy} & \textbf{Communication} \\
& \textbf{Parameter} & \textbf{Time (min)} & \textbf{($\%$)} & \textbf{(Gbits)} \\  
			\hline
			\hline
\textbf{FFT} & $67.0M$ & $3406$& $ 61.6 \pm 1.5 $ & $997$\\
\hline
\textbf{LoRA} & $\mathbf{0.7} M$ & $161$& $50.2 \pm 5$& $\mathbf{45.5}$\\
\hline
\textbf{SFT} & $\mathbf{0.7} M$ & $177$& $55.8\pm 0.1 $& $\mathbf{45.5}$\\
\hline
\textbf{SLoRA} &$\mathbf{0.7} M$  & $\mathbf{160}$ & $\mathbf{56.1 \pm 1}$ & $\mathbf{45.5}$ \\\hline
		\end{tabular}
		\end{adjustbox}
\end{table}

\subsection{Impact of Update size.}
In this section, we show the impact of update size for other settings as well. In all the settings, updates with higher density have better performance as expected, and the performance considerably drops by increasing the data heterogeneity. 

\begin{table}[h]
	\addtolength{\tabcolsep}{-3.5pt}
		\caption{Impact of update density of different PEFT methods on the performance for the 20News group dataset on DistilBERT.}
        \vspace{1mm}
		\label{table:updatesize-2d}
		\centering
		\begin{adjustbox}{width=0.9\linewidth}
		\begin{tabular}{c|c|c|c|c}
 \textbf{Data} & \textbf{Density} & \textbf{LoRA} & \textbf{Holusby} & \textbf{Pfeiffer}\\
\textbf{Distribution}& \textbf{($\%$)} & \textbf{($\%$)} & \textbf{$\%$} & \textbf{($\%$)} \\  \hline \hline
\multirow{2}{*}{Centralized}  & $10$ & $70.0$ & $70.0$ & $69.5$  \\
                              & $1$  & $69.8$ & $69.6$ & $68.7$  \\\hline
\multirow{2}{*}{$\alpha=0.1$}  & $10$ & $66.3$ & $65.3$ & $65.3$  \\
                               & $1$  & $65.8$ & $65.0$ & $64.8$  \\\hline
 \multirow{2}{*}{Pathological}  & $10$ & $58.6$ & $55.1$ & $55.7$  \\
                                 & $1$ &  $57.2$ & $54.7$ & $54.2$ \\\hline
		\end{tabular}
		\end{adjustbox}
\end{table}

\begin{table}[h]
	\addtolength{\tabcolsep}{-3.5pt}
		\caption{Impact of update density of different PEFT methods on the performance for the News category dataset on Albert.}
        \vspace{1mm}
		\label{table:updatesize-ca}
		\centering
		\begin{adjustbox}{width=0.9\linewidth}
		\begin{tabular}{c|c|c|c|c}
 \textbf{Data} & \textbf{Density} & \textbf{LoRA} & \textbf{Holusby} & \textbf{Pfeiffer}\\
\textbf{Distribution}& \textbf{($\%$)} & \textbf{($\%$)} & \textbf{$\%$} & \textbf{($\%$)} \\  \hline \hline
\multirow{2}{*}{Centralized}  & $10$ & $76.0$ & $76.0$ & $76.0$  \\
                              & $1$ & $76.0$ & $75.9$ & $75.9$  \\\hline
\multirow{2}{*}{$\alpha=0.1$}  & $10$ & $72.3$ & $68.54$ & $70.6$  \\
                             & $1$ & $71$ & $68.17$ & $70.27$  \\\hline
 \multirow{2}{*}{Pathological}  & $10$ & $60.7$ & $33.6$ & $57.0$  \\
                               & $1$ & $41.96$ & $27.78$ & $50.8$ \\\hline
		\end{tabular}
		\end{adjustbox}
\end{table}

\begin{table}[h]
	\addtolength{\tabcolsep}{-3.5pt}
		\caption{Impact of update density of different PEFT methods on the performance for the News category dataset on DistilBERT.}
        \vspace{1mm}
		\label{table:updatesize-cd}
		\centering
		\begin{adjustbox}{width=0.9\linewidth}
		\begin{tabular}{c|c|c|c|c}
 \textbf{Data} & \textbf{Density} & \textbf{LoRA} & \textbf{Holusby} & \textbf{Pfeiffer}\\
\textbf{Distribution}& \textbf{($\%$)} & \textbf{($\%$)} & \textbf{$\%$} & \textbf{($\%$)} \\  \hline \hline
\multirow{2}{*}{Centralized}  & $10$ & $78.4$ & $78.0$ & $78.15$  \\
                              & $1$ & $77.99$ & $77.94$ & $77.85$  \\\hline
\multirow{2}{*}{$\alpha=0.1$}  & $10$ & $73.55$ & $72.2$ & $71.6$  \\
                             & $1$ & $71.6$ & $71.6$ & $71.362$  \\\hline
\multirow{2}{*}{Pathological}  & $10$ & $49.39$ & $38.03$ & $37.4$  \\
                               & $1$ & $33.9$ & $33.9$ & $28.2$ \\\hline
		\end{tabular}
		\end{adjustbox}
\end{table}
\end{document}